\documentclass[10pt,twocolumn,letterpaper]{article}

\usepackage{cvpr}
\usepackage{times}
\usepackage{epsfig}
\usepackage{graphicx}
\usepackage{amsmath}
\usepackage{amssymb}
\usepackage{multirow}
\usepackage{booktabs}

\usepackage{algorithm}
\usepackage{algorithmic}

\usepackage{amsmath}

\usepackage[breaklinks=true,bookmarks=false]{hyperref}

\cvprfinalcopy 

\def\cvprPaperID{****} 

\newcommand{\vatex}{\textsc{VaTeX}\xspace}

\begin{document}

\title{Integrating Temporal and Spatial Attentions for  \vatex Video Captioning Challenge 2019}

\author{Shizhe Chen\textsuperscript{1}, Yida Zhao\textsuperscript{1}, Yuqing Song\textsuperscript{1}, Qin Jin\textsuperscript{1},
Qi Wu\textsuperscript{2}\\
\textsuperscript{1} Renmin University of China, \textsuperscript{2} University of Adelaide\\
{\tt\footnotesize \{cszhe1, zyiday, syuqing, qjin\}@ruc.edu.cn, qi.wu01@adelaide.edu.au}
}

\maketitle

\begin{abstract}
This notebook paper presents our model in the \vatex video captioning challenge.
In order to capture multi-level aspects in the video, we propose to integrate both temporal and spatial attentions for video captioning.
The temporal attentive module focuses on global action movements while spatial attentive module enables to describe more fine-grained objects.
Considering these two types of attentive modules are complementary, we thus fuse them via a late fusion strategy.
The proposed model significantly outperforms baselines and achieves 73.4 CIDEr score on the testing set which ranks the second place at the \vatex video captioning challenge leaderboard 2019.
\end{abstract}

\section{Introduction}

Video captioning is a complicated task which requires to recognize multiple semantic aspects in the video such as scenes, objects, actions \etc. and generate a sentence to describe such semantic contents. 
Therefore, it is important to capture multi-level details from global events to local objects in the video in order to generate an accurate and comprehensive video description.
However, state-of-the-art video captioning models \cite{chen2019activitynet,mun2019streamlined} mainly utilize overall video representations or temporal attention \cite{yao2015describing}  over short video segments to generate video description, which lack details at the spatial object level and thus are prone to miss or predict inaccurate details in the video.

In this work, we propose to integrate spatial object level information and temporal level information for video captioning.
The temporal attention is employed to aggregate action movements in the video; while the spatial attention enables the model to ground on fine-grained objects in caption generation.
Since the temporal and spatial information are complementary, we utilize a late fusion strategy to combine captions generated from the two types of attention models.   
Our proposed model achieves consistent improvements over baselines with CIDEr of 88.4 on the \vatex validation set and 73.4 on the testing set, which wins the second place in the \vatex challenge leaderboard 2019.

\section{Video Captioning System}

Our video captioning system consists of three modules, including video encoder to extract global, temporal and spatial features from the video, language decoder that employ spatial and temporal attentions respectively to generate sentences, and temporal-spatial fusion module to integrate generated sentences from different attentive captioning models.

\begin{table*}
	\centering
	\caption{Captioning performance on \vatex validation and testing set.}
	\label{tab:expr_results}
	\begin{tabular}{l|c|cccc|cccc} \toprule
		& & \multicolumn{4}{c|}{Validation} & \multicolumn{4}{c}{Testing} \\
		& RL & BLEU@4 & METEOR & ROUGE & CIDEr & BLEU@4 & METEOR & ROUGE & CIDEr \\ \midrule
		vanilla & & 36.7 & 25.5 & 52.1 & 72.5 & - & - & - & - \\ \midrule
		temporal attention&  & 38.4 & 26.5 & 53.2 & 78.5 & 35.7 & 25.1 & 51.5 & 65.3 \\
		spatial attention & & 38.4 & 26.3 & 53.2 & 77.8 & 36.7 & 25.3 & 52.1 & 66.5 \\ \midrule
		temporal attention & \checkmark & 41.4 & 26.8 & 54.6 & 85.4 & 38.4 & 25.3 & 52.8 & 70.9 \\
		spatial attention & \checkmark & 41.0 & 26.9 & 54.6 & 85.4 & - & - & - & - \\ \midrule
		\textbf{temporal + spatial} & \checkmark & \textbf{42.2} & \textbf{27.4} & \textbf{55.2} & \textbf{88.8} & \textbf{39.1} & \textbf{25.8} & \textbf{53.3} & \textbf{73.4} \\ \bottomrule
	\end{tabular}
\end{table*}

\subsection{Video Encoding}
\paragraph{Global Video Representation.}
In order to comprehensively encode videos as global representation, we extract multi-modal features from three modalities including image, motion and audio.
For the image modality, we utilize the resnext101 \cite{xie2017aggregated} pretrained on the Imagenet to extract global image features every 32 frames and apply average pooling on the temporal dimension; for the motion modality, we utilize the ir-csn model  \cite{tran2019video} pretrained on Kinetics 400 to extract video segment features every 32 frames followed by average pooling; for the audio modality, we utilize VGGish network \cite{hershey2017cnn} pretrained on Youtube8M to extract acoustic features.
We concatenate all the three features and employ linear transformation to obtain the global multi-modal video representation $\bar{v}$.  

\paragraph{Temporal Branch.}
In order to capture action movements, we represent the video as a sequence of consecutive segment-level features in the temporal branch.
Since the image and motion modality features can be well aligned in the temporal dimension, we concatenate the image feature and the motion feature every 32 frames as the segment-level feature, and then apply linear transformation to obtain the fused embedding for each segment as $V^t = \{v^t_i, \cdots, v^t_n\}$.

\paragraph{Spatial Branch.}
In order to capture fine-grained object details, we employ a Mask R-CNN \cite{he2017mask} pretrained on MSCOCO to detect objects in the video every 32 frames.
At most 10 objects are kept for each frame after NMS.
We utilize ROI align \cite{he2017mask} to generate object-level features from feature maps of above mentioned image and motion networks and encode features with linear transformation to generate spatial embeddings as $V^o=\{v^o_1, \cdots, v^o_m\}$.

\subsection{Language Decoding}
We utilize a two-layer LSTM \cite{anderson2018bottom} as the language decoder.
The first layer is an attention LSTM, which aims to generate a query vector $h^1_t$ for attention via collecting necessary contexts from global video representation $\bar{v}$, previous word embedding $w_{t-1}$ and previous output of the second-layer LSTM $h^2_{t-1}$ as follows:
\begin{equation}
h^1_t = \mathrm{LSTM}([\bar{v}; w_{t-1}; h^2_{t-1}], h^1_{t-1})
\end{equation}
Then given the attended memories $V^x, x \in [t, o]$, the query vector $h^1_t$ dynamically fuses relevant features in $V^x$ as $c_t$ via the attention mechanism:
\begin{eqnarray}
&	\alpha_{t, \cdot} = \mathrm{softmax} (w_c^{T} \mathrm{tanh} (W_{vc} v^x_{t,\cdot} + W_{hc} h^1_t)) \\
&	c_t = \sum_i \alpha_{t, i} v^x_i
\end{eqnarray}
The contextual feature $c_t$ will be fed into the second-layer LSTM to predict words where $y_t$ is the target word at the $t$-th step:
\begin{eqnarray}
h^2_t = \mathrm{LSTM}([c_t; h^1_{t}], h^2_{t-1}) \\
p(y_t|y_{<t}) = \mathrm{softmax}(W_y h^2_t + b_y)
\end{eqnarray}
We employ cross entropy loss to train the video encoder and language decoder.
In order to further boost captioning performance in terms of evaluation metrics, we also fine-tune the model with reinforcement learning (RL) \cite{rennie2017self}.
Specifically, we utilize CIDEr and BLEU scores as reward in RL and combine it with cross entropy loss for training.

\subsection{Temporal-Spatial Fusion}
The temporal attentive model and spatial attentive model are complementary with each other since they focus on different aspects in the video.
Therefore, we utilize a late fusion strategy to fuse results from the two types of attentive models.
We train a video-semantic embedding model \cite{faghri2017vse++} on the \vatex dataset, and use it to select the best video descriptions among generated captions from the two types of attentive models according to their relevancy to the video.

\section{Experiments}

Following the challenge policy, we only employ the \vatex dataset \cite{wang2019vatex} for training. 
Since some videos are unavailable to download, our training, validation and testing set contain 25,442, 2,933 and 5,781 videos respectively.
To submit results on the server which requires predictions on full testing set, we further train models using the provided i3d features to generate captions for unavailable videos.

\begin{figure}
	\includegraphics[width=\linewidth]{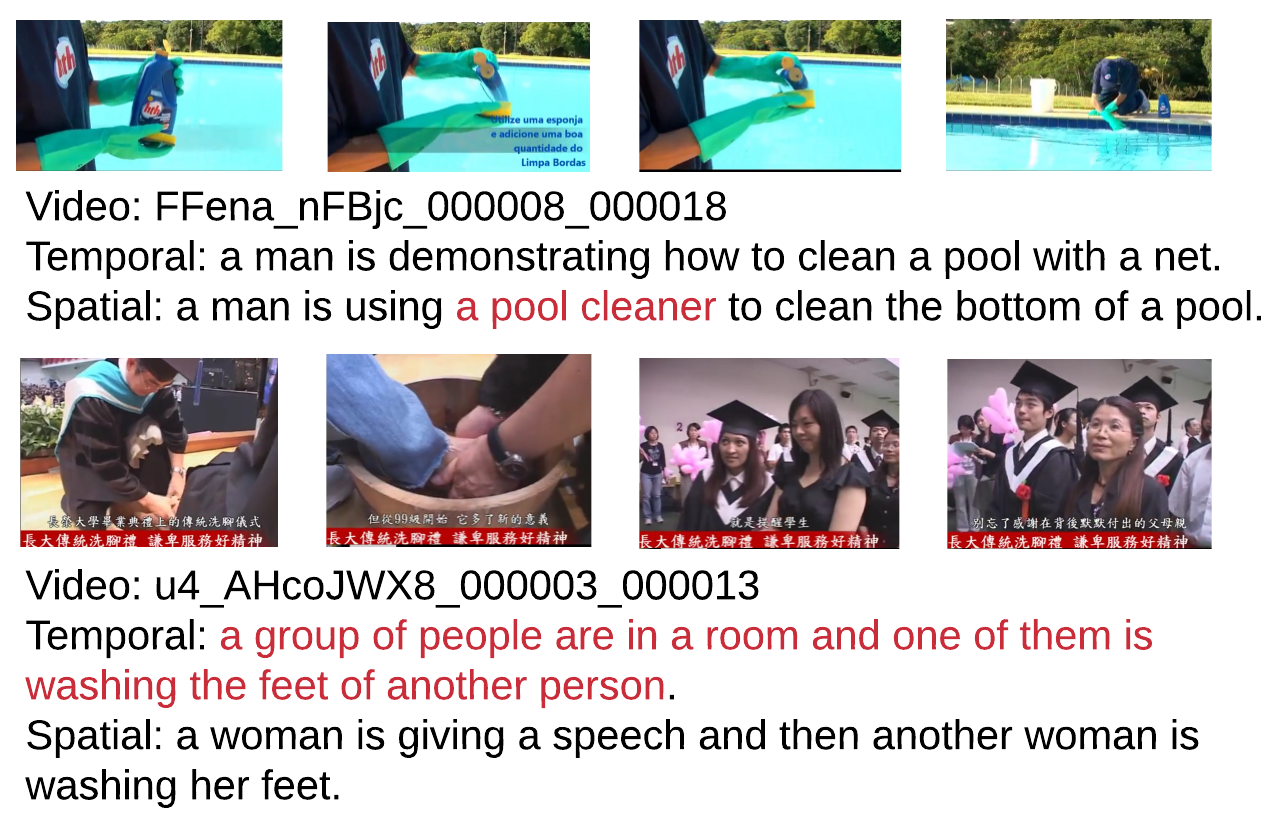}
	\caption{Examples on validation set with generated captions from temporal attentive model and spatial attentive model. The spatial attentive model is better at describing fine-grained object while the temporal attentive model can describe events more accurately.}
	\label{fig:examples}
\end{figure}

Table~\ref{tab:expr_results} represents captioning performance of different captioning models on \vatex dataset.
The vanilla model only utilizes the global video representation without any attention mechanism, which is inferior to temporal and spatial attentive models.
We can see that the temporal and spatial attentions are comparable on the validation set, but the spatial attention achieves slightly better performance on the testing set.
Since the testing videos might contain zero-/few-shot actions, the spatial attention to attend objects might be more generalizable on those videos than temporal attention that focus on action movements.
In Figure~\ref{fig:examples}, we show some generated captions from temporal and spatial models for videos on the validation set.
We can see that the captions from temporal and spatial models are diverse which focus on different aspects of videos, for example, the spatial attentive model can generate descriptions about small objects in the video while the temporal attentive model tends to emphasize on global event.
Therefore, it is beneficial to combine the two types of models.
After fusing temporal and spatial attentive models via our late fusion strategy, we achieve the best performance as shown in Table~\ref{tab:expr_results}.

\section{Conclusion}
In the \vatex Challenge 2019, we mainly focus on integrating temporal and spatial attentions for video captioning, which are complementary for comprehensive video description generation.
In the future, we will explore more effective methods for spatial-temporal reasoning and fusion.
Besides, we will improve the generalization of our models and reduce the performance gap between frequent and few-/zero-shot action videos, such as using stronger video features pretrained on larger dataset like Kinetics 600 and ensemble of different video captioning models.


{\small
\bibliographystyle{unsrt}
\bibliography{reference}
}

\end{document}